    \documentclass[letterpaper, 10 pt, conference]{ieeeconf}  

    \IEEEoverridecommandlockouts                              
    
    \usepackage{graphicx}
    \usepackage{tikz}
    \usepackage{caption}
    \usepackage{subcaption}
    \usepackage{color}
    \usetikzlibrary{arrows}
    \usetikzlibrary{calc}
    \usepackage{cite}
    \usepackage{amsmath,amssymb,amsfonts}
    \usepackage{algorithmic}
    \usepackage{graphicx}
    \usepackage{textcomp}
    \usepackage{xcolor}
    \usepackage{multirow}
    
    \usepackage{amssymb}
    \usepackage{pifont}
    \newcommand{\cmark}{\ding{51}}%
    \newcommand{\xmark}{\ding{55}}%
  
  \newcommand\blfootnote[1]{%
  \begingroup
  \renewcommand\thefootnote{}\footnote{#1}%
  \addtocounter{footnote}{-1}%
  \endgroup
}  

\def\footnoterule{\relax%
  \kern-5pt
  \hbox to \columnwidth{\vrule width .4\columnwidth height 0.4pt\hfill}
  \kern4.6pt}
\makeatother

    \title{\LARGE \bf
    Towards Real-time Traffic Sign and Traffic Light Detection on Embedded Systems}

    \author{Oshada Jayasinghe, Sahan Hemachandra, Damith Anhettigama, Shenali Kariyawasam, 
    Tharindu \\ Wickremasinghe, Chalani Ekanayake, Ranga Rodrigo and Peshala Jayasekara
    \thanks{Department of Electronic and Telecommunication Engineering, University of Moratuwa, Sri Lanka}}

\begin{document}    

    \maketitle
    \thispagestyle{plain}
    \pagestyle{plain}

    \begin{abstract}
    
    Recent work done on traffic sign and traffic light detection focus on improving detection accuracy in complex scenarios, yet many fail to deliver real-time performance, specifically with limited computational resources. In this work, we propose a simple deep learning based end-to-end detection framework, which effectively tackles challenges inherent to traffic sign and traffic light detection such as small size, large number of classes and complex road scenarios. We optimize the detection models using TensorRT and integrate with Robot Operating System to deploy on an Nvidia Jetson AGX Xavier as our embedded device. The overall system achieves a high inference speed of 63 frames per second, demonstrating the capability of our system to perform in real-time. Furthermore, we introduce CeyRo, which is the first ever large-scale traffic sign and traffic light detection dataset for the Sri Lankan context. Our dataset consists of 7984 total images with 10176 traffic sign and traffic light instances covering 70 traffic sign and 5 traffic light classes. The images have a high resolution of 1920 x 1080 and capture a wide range of challenging road scenarios with different weather and lighting conditions. Our work is publicly available at \url{https://github.com/oshadajay/CeyRo}.
    

    \end{abstract}
    
    \blfootnote{\textcopyright \ 2022 IEEE. Personal use of this material is permitted.
  Permission from IEEE must be obtained for all other uses, in any current or future 
  media, including reprinting/republishing this material for advertising or promotional 
  purposes, creating new collective works, for resale or redistribution to servers or 
  lists, or reuse of any copyrighted component of this work in other works.}

    
    \section{Introduction}
    
    Traffic signs and traffic lights play a vital role in regulating the traffic by providing necessary information to drivers to safely maneuver on roads. Thus detection of these two elements becomes a fundamental perception task involved in the development of autonomous vehicles and advanced driver assistance systems (ADAS). Developing robust detection algorithms can be a challenging task due to several reasons. Traffic signs and traffic lights usually occupy a small area of a typical street view image. It can be hard to differentiate traffic signs from other similar objects such as billboards and advertisement boards. The algorithms should be robust to occlusions, illumination changes, varying weather conditions and the deterioration of traffic signs with time. While addressing these challenges, it is essential for a detection system to deliver real-time performance with limited computational resources.

    
    
    Initial work done on traffic sign and traffic light detection \cite{tfs_overett2011, banes2015semantic, maldonado2007, tfs_ellahyani2016, gomez2014traffic, jenson2015traffic} mainly focus on traditional image processing based techniques and machine learning based algorithms. Recent deep learning based approaches \cite{tfs_TT100K, tfs_segunet, tfs_largescale, behrendt2017deep, bach18, tfs_GAN} have been able to outperform these classical approaches, especially in challenging and complex road scenarios. However, most of these implementations are carried out on high-end graphics processing units (GPUs) and less attention is given to delivering real-time performance on embedded systems, which is crucial for both autonomous vehicles and ADAS implementations.
    
    In this work, we propose an end-to-end, deep learning based traffic sign and traffic light detection framework, which is robust to challenging road scenarios, and demonstrates real-time performance on an embedded system. We first create the CeyRo traffic sign and traffic light dataset consisting of 7984 total images belonging to 70 traffic sign and 5 traffic light classes. Our dataset comprises 10176 traffic sign and traffic light instances and covers a wide variety of challenging urban, sub-urban and rural road scenarios. 
    
    Our traffic sign and traffic light detection pipeline consists of two stages: 1. Detection or localization of the traffic sign or the traffic light in the original image considering the superclass. 2. Classification of each detection to its respective class. We train and evaluate the performance of two state-of-the-art object detectors, Faster R-CNN \cite{faster_rcnn} and SSD \cite{SSD} for the detection task and a separate ResNet-18 \cite{resnet} classifier is trained for the classification task.
    
    
        \begin{figure*}[h]
        \begin{center}
            \includegraphics[width=0.98\linewidth]{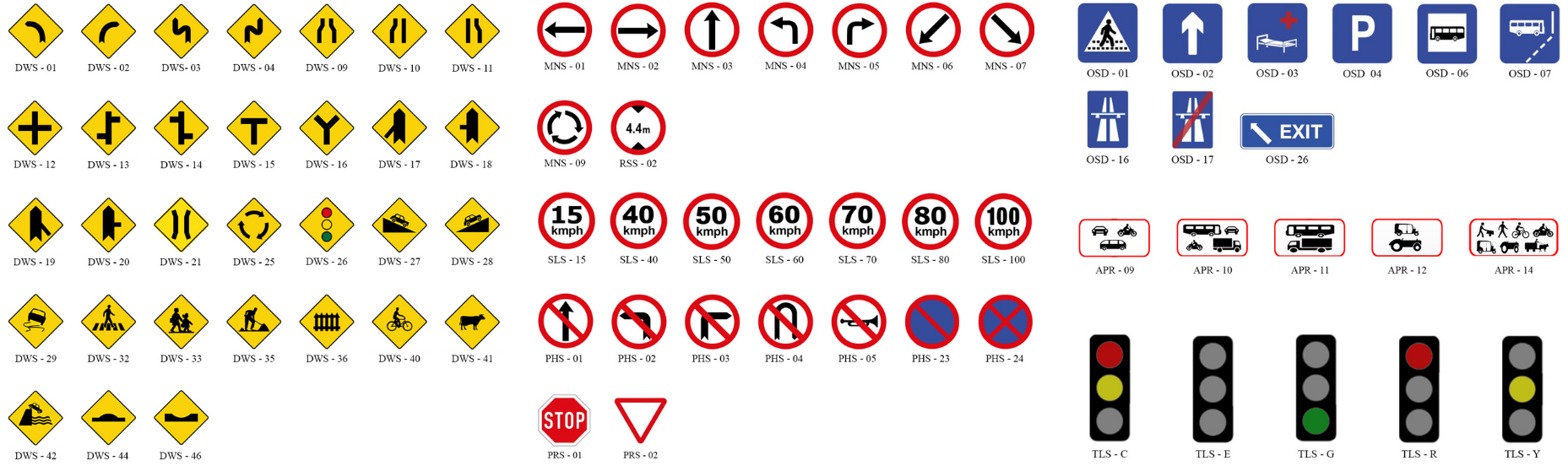}
             \caption{The 70 traffic sign classes and the 5 traffic light classes of the CeyRo dataset.}
            
            \label{fi:ts_tl_grid}
        \end{center}
    \end{figure*}
    
    
    We then optimize both our detector and classifier models using TensorRT and integrate with Robot Operating System (ROS) \cite{ros} to deploy as a traffic sign and traffic light detection system on an Nvidia Jetson AGX Xavier device. The overall system is well capable of delivering real-time performance with a high inference speed of 63 frames per second (FPS). In summary, our contributions are three-fold:
    
    \begin{itemize}
        \item We introduce CeyRo: the first ever large-scale dataset for traffic sign and traffic light detection within the Sri Lankan context.
        
        
        \item We evaluate the approach of utilizing a two-staged convolutional neural network based model architecture for the traffic sign and traffic light detection task, and provide results in terms of speed and accuracy.
        
        \item We demonstrate the capability of our detection framework to perform in real-time, by deploying the trained and optimized models on an embedded system, integrated with the ROS ecosystem.
    \end{itemize}

    \section{Related Work}

            
    
    

    Traditional image processing based algorithms, as well as deep learning based approaches, have been used for both traffic sign detection and traffic light detection tasks. The availability of large-scale, high quality visual datasets has become a critical factor for the development of these algorithms, especially with deep learning. A summary of widely used publicly available traffic sign detection and traffic light detection datasets are presented in Table \ref{tab:tsdatasets} and Table \ref{tab:tldatasets}, respectively.
    
    \begin{table}[t]
    \small
    \begin{center}
    \caption{Traffic sign detection benchmark datasets.}
    \label{tab:tsdatasets}
    \begin{tabular}{|l|c|c|c|c|c|}
    \hline
    \textbf{Dataset} & \textbf{Year} & \textbf{Annotations} & \textbf{Classes} & \textbf{Location} \\
    \hline
    LISA \cite{LISA} & 2012 & 7855 & 47 & USA \\
    \hline
    GTSDB \cite{GTSDB} & 2013 & 852 & 43 & Germany  \\
    \hline
    TT100K \cite{tfs_TT100K} & 2016  & 26349 & 221 & China   \\
    \hline
    MTSD \cite{MTSD} & 2020 & 325172 & 313 & Worldwide  \\
    \hline
    
    \end{tabular}
    \end{center}
    
    \end{table}

    Traditional image processing based approaches like colour, ratio and shape based filtering and hand-craft feature extraction have been used in \cite{tfs_overett2011} and \cite{tfs_ellahyani2016} for traffic sign detection and recognition. Similarly, for traffic lights, distinct features of traffic lights such as colour and shape have been used in \cite{gomez2014traffic}, while \cite{banes2015semantic} uses HOG features. Machine learning based techniques such as support vector machines (SVMs) and random forests have been used for traffic sign classification in \cite{maldonado2007} and \cite{tfs_ellahyani2016}. Hidden Markov models and SVMs have been used as machine learning based techniques for traffic light detection and classification in \cite{gomez2014traffic} and \cite{jenson2015traffic}. Even though these methods are less computationally complex, they have limited usage scenarios and do not perform well in challenging environments when compared with recent deep learning based approaches.
    
    A robust end-to-end convolutional neural network is proposed in \cite{tfs_TT100K}, which outperforms state-of-the-art object detectors for their TT100K dataset, particularly with small traffic sign detection. The traffic sign detection and recognition problem becomes complex with the increased number of classes. In \cite{tfs_largescale}, this problem is addressed to detect and recognize around 200 traffic sign classes following the Mask R-CNN \cite{mask-rcnn} architecture. A semantic segmentation based approach is followed in Seg-U-NET \cite{tfs_segunet}, where two state-of-the-art segmentation architectures are combined for detecting traffic signs and a separate classifier is used for the recognition part. A perceptual generative adversarial network (GAN), which consists of a generator network and a discriminator network, is used in \cite{tfs_GAN} to detect small traffic signs with higher accuracy. Considering deep learning based approaches for traffic light detection, \cite{behrendt2017deep} has introduced a YOLO \cite{yolo_cvpr_2016} based detection network and a small classification network to accurately detect and classify traffic lights. A Fast R-CNN \cite{fast_rcnn_iccv_2015} based network architecture has been used in \cite{bach18} for traffic light detection with the DriveU \cite{DriveU} dataset. 
    
    \begin{table}[t]
    \small
    \begin{center}
    \caption{Traffic light detection benchmark datasets.}
    \label{tab:tldatasets}
    \begin{tabular}{|l|c|c|c|c|c|}
    \hline
    \textbf{Dataset} & \textbf{Year} & \textbf{Annotations} & \textbf{Classes} & \textbf{Location}\\
    \hline
    Bosch \cite{behrendt2017deep} & 2017 & 24242 & 15 & Germany  \\
    \hline
    DriveU \cite{DriveU} & 2018 & 232039 & 344 & Germany \\
    \hline
    \end{tabular}
    \end{center}
    
    \end{table}
    
    

     \begin{figure*}[t]
        \begin{center}
            \input{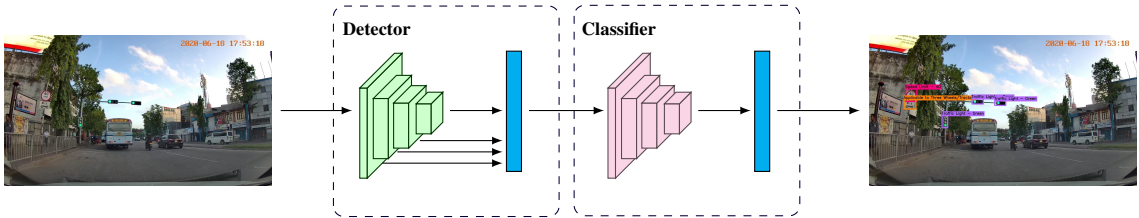}
            \caption{Traffic sign and traffic light detection model architecture. Faster-RCNN-ResNet50 \cite{faster_rcnn,resnet} and SSD-MobileNet-v2 \cite{SSD,sandler2018mobilenetv2} are evaluated as object detectors to first detect the traffic signs and traffic lights as bounding boxes under their superclasses. Then a ResNet-18 \cite{resnet} classifier is used to classify the detections into their respective classes.}
            \label{fi:ts_tl}
        \end{center}
    \end{figure*}

    High-end computational platforms have been used for almost all of the above mentioned algorithms, and implementation of traffic sign and traffic light detection systems on embedded systems is not a widely researched field. A colour based detection algorithm has been used in \cite{roadsignfpga} to identify regions of interest and classify road signs into four categories using a Xlinix Spartan-3A DSP FPGA device. An Artix-7 FPGA device is used in \cite{tl_fpga} to implement a parallelism optimized AdaBoost based detection algorithm for real-time traffic light detection. Both of these approaches rely on classical techniques, which often have limited usage scenarios when compared with modern deep learning based approaches. Nvidia Jetson TX1 and TX2 devices are used in \cite{ouyang19} for traffic light detection, which achieves an inference speed of 10 FPS. Their approach also uses a heuristic colour based candidate region selection module for identifying regions of interest, and a lightweight convolution neural network (CNN) has been used only for the classification part.

    \section{Benchmark Dataset}
    
    \subsection{Data Collection}
    
    Most of the publicly available traffic sign and traffic light datasets are created using footage captured from cameras mounted on vehicles \cite{GTSDB, behrendt2017deep} or images extracted from street view services such as Google or Tencent \cite{tfs_TT100K}. For the creation of our dataset, we collect video footage from two cameras mounted inside four vehicles and the frames containing traffic signs or traffic lights are manually extracted. We cover a wide range of challenging scenarios in urban, rural and expressway conditions in Sri Lanka, which include different weather and lighting conditions, occlusions and deteriorated signs.

    \subsection{Data Annotation}
    
    We use the LabelImg \cite{labelImg} image annotation tool to manually annotate the traffic signs and traffic lights present in the extracted images as bounding boxes. For each image, an XML file is created in the PASCAL VOC \cite{pascalvoc} format containing the bounding box annotations of the traffic sign and traffic light instances.
    
    \subsection{Dataset Statistics}
    
    Our benchmark dataset consists of 7984 total images with a resolution of $1920 \times 1080$. The dataset is divided into the train set and the test set, comprising 6143 images and 1841 images, respectively. The dataset covers 70 different traffic sign classes and 5 traffic light classes which are visualized in Fig. \ref{fi:ts_tl_grid}. There is an inherent class imbalance in the dataset since some traffic sign and traffic light classes are not found very often. Classes which have less than 25 total instances have been excluded from the test set. The traffic sign and traffic light classes can be further categorized into 8 superclasses and the number of instances present in each superclass is shown in Table \ref{tab:tl_superclass}.
    
        \begin{table}[t]
    \vspace{0.14cm}
    \caption{Number of instances for each superclass in the CeyRo dataset.}
\begin{center}
\begin{tabular}{|l|c|c|c|}
\hline
\textbf{Superclass} & \textbf{Train} & \textbf{Test} & \textbf{Total} \\
\hline\hline
Danger Warning Signs (DWS)	& 2833 & 809 & 3642\\
\hline
Mandatory Signs	(MNS) & 453 &	128 & 581\\
\hline
Prohibitory Signs (PHS) & 650 &	195 & 845\\
\hline
Priority Signs (PRS) & 115 & 26 & 141\\
\hline
Speed Limit Signs (SLS) & 735 &	237 & 972\\
\hline
Other Signs Useful for Drivers (OSD) & 1619 &	498 & 2117\\
\hline
Additional Regulatory Signs (APR) & 377 &	123 & 500\\
\hline
Traffic Light Signs (TLS) & 1075 &	303 & 1378\\
 \hline 
\textbf{Total} & \textbf{7857} & \textbf{2319} & \textbf{10176}\\
\hline
\end{tabular}
\end{center}



        \label{tab:tl_superclass}
    \end{table}
    
    

    \subsection{Evaluation Metric}
    
    We use $F_{1}$-score as the evaluation metric of our traffic sign and traffic light dataset. Each prediction with an intersection over union (IoU) higher than 0.3 with the ground truth is considered as a true positive. The precision, recall and $F_{1}$-score can be then calculated as follows where $TP$, $FP$, $FN$ denotes the total number of true positives, false positives and false negatives, respectively. 
    \begin{equation}
    precision = \frac{TP}{TP+FP}
    \end{equation}
    
    \begin{equation}
    recall = \frac{TP}{TP+FN}
    \end{equation}
    
    \begin{equation}
        F_{1}\mbox{-}score= \frac{2\times precision \times recall}{precision+recall}
    \end{equation}
    
    \vspace{0.2em}
    
    \section{Methodology}
    
    \begin{figure*}[h]
        \begin{center}
            \begin{tikzpicture}

\setlength{\baselineskip}{5em}

\tikzstyle{bluebox}=[draw=cyan!20!black,fill=cyan!20]
\tikzstyle{greenbox}=[draw=green!20!black,fill=green!20]
\tikzstyle{orangebox}=[draw=yellow!20!black,fill=yellow!20]
\tikzstyle{magentabox}=[draw=magenta!20!black,fill=magenta!20]

\tikzstyle{labelnode}=[align=center, execute at begin node=\setlength{\baselineskip}{0.75em}]

\newcommand{\cube}[5][]
{
	\pgfmathsetmacro{\cubex}{(#2)}
	\pgfmathsetmacro{\cubey}{{#3}}
	\pgfmathsetmacro{\cubez}{{#4}}
	\draw[#5] (0,0,0) -- ++(-\cubex,0,0) -- ++(0,-\cubey,0) -- ++(\cubex,0,0) -- cycle;
	\draw[#5] (0,0,0) -- ++(0,0,-\cubez) -- ++(0,-\cubey,0) -- ++(0,0,\cubez) -- cycle;
	\draw[#5] (0,0,0) -- ++(-\cubex,0,0) -- ++(0,0,-\cubez) -- ++(\cubex,0,0) -- cycle;
}

\begin{scope}[xshift=0cm, yshift=0cm]
    \node (pf) at (-.4,.7) [anchor=west, draw=black, text width=2.2cm, align=center, inner sep=5pt, labelnode,fill=green!20] {\scriptsize Frozen Inference Graph};
    \node (pf) at (-.4,-.7) [anchor=west, draw=black, text width=2.2cm, align=center, inner sep=5pt, labelnode,fill=green!20] {\scriptsize Configuration Parameters};
    \draw [-latex] (2.3, .7) -- ++(.6, -.6);
    \draw [-latex] (2.3, -.7) -- ++(.6, +.6);
\end{scope}

\begin{scope}[xshift=3cm, yshift=0cm]
    \node (pf) at (0,0) [anchor=west, draw=black, text width=1.8cm, align=center, inner sep=5pt, labelnode,fill=cyan!20] {\scriptsize UFF Library \& GraphSurgeon};
    \draw [-latex] (2.25,0) -- ++(.65, 0);
    \node (pf) at (3,0) [anchor=west, draw=black, text width=1.8cm, align=center, inner sep=5pt, labelnode,fill=green!20] {\scriptsize Intermediate UFF File};
    \draw [-latex] (5.25,0) -- ++(.55, 0);
    \node (pf) at (5.9,0) [anchor=west, draw=black, text width=1.8cm, align=center, inner sep=5pt, labelnode,fill=cyan!20] {\scriptsize TensorRT UFF Parser};
    \draw [-latex] (8.15,0) -- ++(.75, 0);
    \node (pf) at (9,0) [anchor=west, draw=black, text width=1.8cm, align=center, inner sep=5pt, labelnode,fill=green!20] {\scriptsize  FP16 TensorRT Engine};
\end{scope}

\end{tikzpicture}
            \caption{TensorRT conversion and quantization process of the detector model trained using TensorFlow Object Detection API.}
            
            \label{fi:opt_detector}
        \end{center}
    \end{figure*}
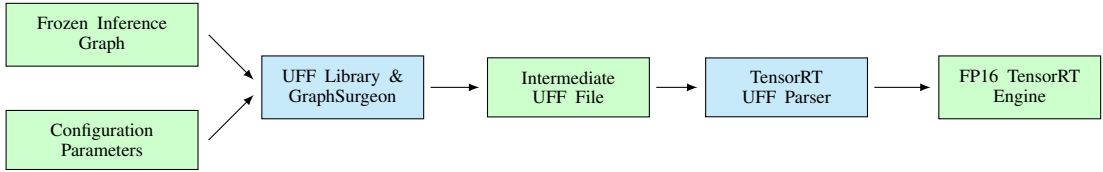
    
    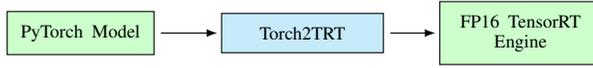
\begin{figure*}[h]
        \begin{center}
            \begin{tikzpicture}

\setlength{\baselineskip}{5em}

\tikzstyle{bluebox}=[draw=cyan!20!black,fill=cyan!20]
\tikzstyle{greenbox}=[draw=green!20!black,fill=green!20]
\tikzstyle{orangebox}=[draw=yellow!20!black,fill=yellow!20]
\tikzstyle{magentabox}=[draw=magenta!20!black,fill=magenta!20]

\tikzstyle{labelnode}=[align=center, execute at begin node=\setlength{\baselineskip}{0.75em}]

\newcommand{\cube}[5][]
{
	\pgfmathsetmacro{\cubex}{(#2)}
	\pgfmathsetmacro{\cubey}{{#3}}
	\pgfmathsetmacro{\cubez}{{#4}}
	\draw[#5] (0,0,0) -- ++(-\cubex,0,0) -- ++(0,-\cubey,0) -- ++(\cubex,0,0) -- cycle;
	\draw[#5] (0,0,0) -- ++(0,0,-\cubez) -- ++(0,-\cubey,0) -- ++(0,0,\cubez) -- cycle;
	\draw[#5] (0,0,0) -- ++(-\cubex,0,0) -- ++(0,0,-\cubez) -- ++(\cubex,0,0) -- cycle;
}

\begin{scope}[xshift=3cm, yshift=0cm]
    \node (pf) at (0.15,0) [anchor=west, draw=black, text width=1.6cm, align=center, inner sep=5pt, labelnode,fill=green!20] {\scriptsize PyTorch Model};
    \draw [-latex] (2.2,0) -- ++(.75, 0);
    \node (pf) at (3,0) [anchor=west, draw=black, text width=1.8cm, align=center, inner sep=5pt, labelnode,fill=cyan!20] {\scriptsize Torch2TRT};
    \draw [-latex] (5.25,0) -- ++(.6, 0);
    \node (pf) at (5.9,0) [anchor=west, draw=black, text width=1.8cm, align=center, inner sep=5pt, labelnode,fill=green!20] {\scriptsize FP16 TensorRT Engine};
\end{scope}

\end{tikzpicture}
            \caption{TensorRT conversion and quantization process of the classifier model trained using PyTorch.}
            
            \label{fi:opt_classifier}
        \end{center}
    \end{figure*}
    
    \begin{figure*}[h]
        \begin{center}
            \includegraphics[width=0.98\linewidth]{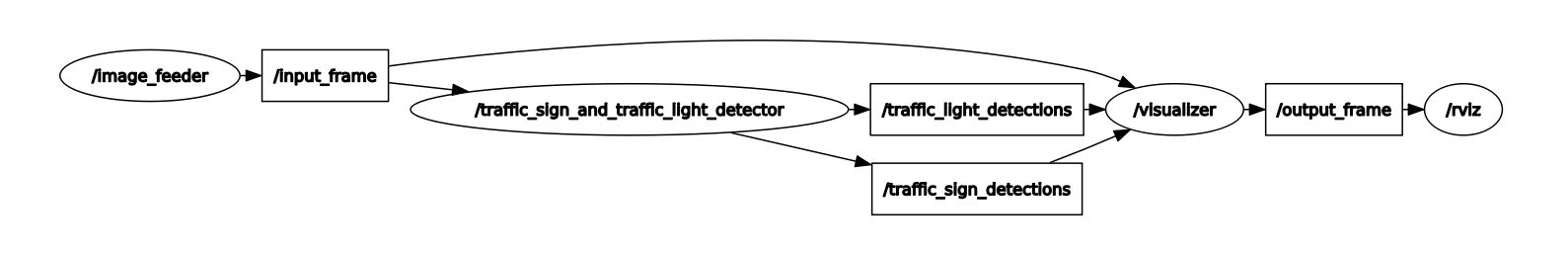}
            \caption{RQT graph for the implementation of the traffic sign and traffic light detection system in the ROS ecosystem.}
            
            \label{fi:rqt}
        \end{center}
    \end{figure*}

    \subsection{Model Architecture}
    
    The proposed traffic sign and traffic light detection model architecture is shown in Fig. \ref{fi:ts_tl}. A state-of-the-art object detector model is used to first detect the traffic signs and traffic lights present in the input image as bounding boxes under their 8 superclasses. Then a separate classifier model is used to classify each detection into its respective class.
     
     
     We evaluate the performance of two state-of-the-art object detectors for the traffic sign and traffic light detection task. Faster R-CNN \cite{faster_rcnn} is used as a two stage object detector and SSD \cite{SSD} is used as a single stage object detector. ResNet-50 \cite{resnet} is used as the backbone of the Faster R-CNN model while MoblieNet-v2 \cite{mobilenet} is used as the backbone of the SSD model. The input resolution is set to $ 512 \times 512 $ in both object detector models.
     
     A ResNet-18 \cite{resnet} classifier is trained for the traffic sign and traffic light classification task. The input image resolution is set to $ 100 \times 100 $ and the number of output classes is set to 75, which includes the 70 traffic sign classes and the 5 traffic light classes. The traffic sign and traffic light instances present in the train set of our dataset are cropped and extracted out to create the train set for the classifier.
     
     
            

    
    \subsection{Model Training}
    
    We use TensorFlow Object Detection API \cite{huang2017speed} to train the two object detector models. For training the SSD-MobileNet-v2 \cite{SSD,sandler2018mobilenetv2} model, RMSProp \cite{ruder2016overview} optimization is used with an initial learning rate of 0.004 and a momentum of 0.9, and the batch size is set to 24. For training the Faster-RCNN-ResNet50 \cite{faster_rcnn,resnet} model, SGD with momentum \cite{pmlr-v28-sutskever13} optimization is used with an initial learning rate of 0.0003 and a momentum of 0.9, and the batch size is set to 8. 
    
    The ResNet-18 \cite{resnet} classifier is trained for 30 epochs using PyTorch \cite{paszke2017automatic}. The cross entropy loss is used as the loss function and SGD algorithm with a learning rate of 0.01 and a momentum of 0.9 is used as the optimization function. The batch size is set to 512. We use a computational platform with an Intel Core i9-9900K CPU and an Nvidia RTX-2080 Ti GPU to train our models.
    
    
    \begin{table*}[t]
    \caption{Traffic sign and traffic light detection results. For each of the detection models, class-wise $F_{1}$-scores, overall precision, overall recall, overall $F_{1}$-score and the inference speed in frames per second (FPS) are listed.}
\begin{center}
\begin{subtable}[t]{0.32\textwidth}
\begin{center}
\begin{tabular}{|l|c|c|}
\hline
\textbf{Label} & \textbf{SSD-512} & \textbf{FRCNN-512}\\
\hline\hline
DWS-01 & 0.9583 & 0.9863\\
\hline
DWS-02 & 0.9774& 0.9778\\
\hline
DWS-03 & 0.9737 & 1.0000\\
\hline
DWS-04 & 1.0000 & 0.9895\\
\hline
DWS-10 & 0.9630 & 0.9434\\
\hline
DWS-11 & 1.0000 & 0.9412\\
\hline
DWS-12 & 1.0000 & 0.9444\\
\hline
DWS-13 & 1.0000 & 0.9412\\
\hline
DWS-14 & 1.0000 & 1.0000\\
\hline
DWS-17 & 0.9765 & 0.9827\\
\hline
DWS-18 & 0.9412 &0.9412\\
\hline
DWS-19 & 0.9412 & 0.9346\\
\hline
DWS-21 & 0.9545 & 0.9778\\
\hline
DWS-25 & 0.9474 & 0.8421\\
\hline
DWS-26 & 0.9825 & 0.9483\\
\hline

DWS-32 & 0.9677 & 0.9579 \\
\hline
DWS-33 & 0.9573 & 0.9836\\
\hline

DWS-35 & 0.9677 & 0.8387\\
\hline
DWS-40 & 1.0000 & 0.9744\\
\hline
DWS-41 & 0.9630 & 1.0000\\
\hline
MNS-01 & 0.8571 & 0.8000\\
\hline
\end{tabular}
\end{center}
\end{subtable}
\begin{subtable}[t]{0.32\textwidth}
\begin{center}
\begin{tabular}{|l|c|c|}
\hline
\textbf{Label} & \textbf{SSD-512} & \textbf{FRCNN-512}\\
\hline\hline
MNS-06 &  0.9529 & 0.9302\\
\hline
MNS-07 &  0.8571 & 0.8696\\
\hline
MNS-09 & 0.9091 & 0.8571\\
\hline
OSD-01 & 0.9394 & 0.8502\\
\hline
OSD-02 & 0.7273 & 0.5714\\
\hline
OSD-03 & 1.0000 & 0.9524\\
\hline
OSD-04 & 0.8400 & 0.8679\\
\hline
OSD-06 & 0.9845 & 0.9731\\
\hline
OSD-07 & 0.9111 & 0.8542\\
\hline
OSD-16 & 0.8696 & 0.8462\\
\hline
OSD-17 & 0.8000 &  0.8421\\
\hline
OSD-26 & 0.9524 & 0.8205\\
\hline
PHS-01 & 0.9268 & 0.8421\\
\hline
PHS-02 & 0.9600 & 0.8800\\
\hline
PHS-03 & 0.9412 & 0.8846\\
\hline
PHS-04 & 0.8679 & 0.8679\\
\hline
PHS-09 & 0.9600 & 0.9630\\
\hline
PHS-23 & 0.9587 & 0.9000\\
\hline
PHS-24 & 0.9355 & 0.9524\\
\hline
PRS-01 & 0.8444 & 0.7451\\
\hline
RSS-02 & 0.9091 & 1.0000\\
\hline
\end{tabular}
\end{center}
\end{subtable}
\begin{subtable}[t]{0.32\textwidth}
\begin{center}
\begin{tabular}{|l|c|c|}
\hline
\textbf{Label} & \textbf{SSD-512} & \textbf{FRCNN-512}\\
\hline\hline
SLS-100 & 0.9706 & 1.0000\\
\hline
SLS-15 & 0.8421 & 0.8571\\
\hline
SLS-40 & 0.9242 & 0.9624\\
\hline
SLS-50 & 0.8684 & 0.8608\\
\hline
SLS-60 & 0.8889 & 0.8889\\
\hline
SLS-70 & 0.9615 & 0.9600\\
\hline
SLS-80  & 0.9333 & 0.9492\\
\hline
APR-09 & 0.9032 & 0.7931\\
\hline
APR-10 & 0.9091 & 0.7692\\
\hline
APR-11 & 0.8824 & 0.7742\\
\hline
APR-12 & 0.8919 & 0.7761\\
\hline
APR-14 & 0.9286 & 0.8571\\
\hline
TLS-C  & 0.6154 & 0.4000 \\
\hline
TLS-E  & 0.5714 & 0.4421\\
\hline
TLS-G & 0.8176 & 0.7673\\
\hline
TLS-R &  0.7407 & 0.7143\\
\hline
TLS-Y  &  0.6769 & 0.7576\\
\hline
\textbf{Precision}  & \textbf{0.9670} & 0.9259\\
\hline
\textbf{Recall}    & \textbf{0.8848} &  0.8676\\
\hline
\textbf{F1-score}   & \textbf{0.9241} & 0.8958\\
\hline
\textbf{FPS} & \textbf{83} & 34\\
\hline
\end{tabular}
\end{center}
\end{subtable}
\end{center}

        \label{ta:results_ts}
    \end{table*}
    
    \subsection{Data Augmentation}
    
    To reduce the effect of the class imbalance problem, we use data augmentation techniques to increase the number of instances of less frequent traffic signs and traffic lights. Random horizontal flip is used as an augmentation technique when training both detector and classifier models to mirror the traffic signs and traffic lights and create new instances where applicable. Furthermore, colour jitter augmentation technique is used when training the classifier to randomly change the brightness, contrast, saturation and hue of the input images.
    
            
            
    
    \subsection{Embedded System Implementation}
    
    We use an Nvidia Jetson AGX Xavier as the embedded device to deploy our traffic sign and traffic light detection system. The detector and classifier models have comparatively low speeds when directly inferenced on the embedded device due to its resource constrained nature. Thus, we optimize the trained models using TensorRT optimization with half-precision floating-point (FP16) quantization to effectively utilize the CUDA and Tensor cores present in the device. We use the SSD-MobileNet-v2 \cite{SSD,sandler2018mobilenetv2} model as the detector for the embedded system implementation, since it is much faster than the Faster-RCNN-ResNet50 \cite{faster_rcnn,resnet} model.
    
    The traffic sign and traffic light detection model trained using the TensorFlow Object Detection API \cite{huang2017speed} can be optimized using TensorRT as shown in Fig. \ref{fi:opt_detector}. First, the frozen inference graph and the configuration parameters of the model are used to generate an intermediate file in the UFF format using graphsurgeon and UFF libraries. Second, the intermediate file is quantized into a FP16 TensorRT engine using the UFF Parser in the TensorRT Python API. The classifier model which was trained using PyTorch \cite{paszke2017automatic} can be optimized using torch2trt \cite{torch2trt} as shown in Fig. \ref{fi:opt_classifier}. A direct conversion of the trained PyTorch model in .pth file format to a FP16 quantized TensorRT engine is facilitated by torch2trt which utilizes the TensorRT Python API.
    
    We implement our traffic sign and traffic light detection system in the Robot Operating System (ROS) \cite{ros} ecosystem as shown in Fig. \ref{fi:rqt}. The \textit{image\_feeder} node retrieves each frame from a given video file and publishes them to the \textit{input\_frame} topic. The \textit{traffic\_sign\_and\_traffic\_light\_detector} node detects traffic signs and traffic lights in the current frame using the generated TensorRT engines. The detections are then published to the \textit{traffic\_sign\_detections} and \textit{traffic\_light\_detections} topics respectively. The \textit{visualizer} node marks the detected traffic signs and traffic lights in the current frame and the resultant image is published to the \textit{output\_frame} topic. The RViz visualization tool can be used to visualize the traffic sign and traffic light detections in real-time.
    
    \section{Results}
    
    The traffic sign and traffic light detection results of the two trained models are tabulated in Table \ref{ta:results_ts}, including the $F_{1}$-scores for the 59 classes in the test set, overall precision, overall recall, overall $F_{1}$-score and the inference speed on the workstation with the Nvidia RTX-2080 Ti GPU. The inference speed is calculated as the average FPS for the 1841 test images. 
    
    It can be observed that the SSD-MobileNet-v2 \cite{SSD,sandler2018mobilenetv2} model detects traffic signs and traffic lights more accurately than the Faster-RCNN-ResNet50 \cite{faster_rcnn,resnet} model. This is contrary to the general belief that two stage object detectors perform better than single stage object detectors. The SSD-MobileNet-v2 \cite{SSD,sandler2018mobilenetv2} model also achieves a higher inference speed of 83 FPS than the Faster-RCNN-ResNet50 \cite{faster_rcnn,resnet} model.
    $F_{1}$-score values for some traffic sign and traffic light classes are comparatively low, which could be mainly due to the lower number of instances of those classes in the train set. 
    
    
    



    The results of the TensorRT optimization process are shown in Table \ref{ta:opt}. Each row indicates whether the detector model is optimized, whether the classifier model is optimized and the resulting $F_{1}$-score and the inference speed on the Nvidia Jetson AGX Xavier device. It can be observed that with a slight drop in accuracy, the inference speed can be increased significantly by optimizing and quantizing both detector and classifier models through TensorRT. 
    
    \begin{table}[t]
    \vspace{0.14cm}
    \caption{TensorRT optimization results. For each combination, resulting $F_{1}$-score and the inference speed are listed.}
        \begin{center}
\begin{tabular}{|c|c|c|c|}
\hline
\multicolumn{2}{|c|}{\textbf{Optimization}}          & \multicolumn{1}{c|}{\multirow{2}{*}{\hspace{0.5em} \textbf{F1-score} \hspace{0.5em}}} & \multicolumn{1}{c|}{\multirow{2}{*}{\hspace{0.5em} \textbf{FPS} \hspace{0.5em}}} \\ \cline{1-2}
\multicolumn{1}{|l|}{\hspace{1em} \textbf{Detector} \hspace{1em} } & \hspace{1em} \textbf{Classifier} \hspace{1em}  & \multicolumn{1}{c|}{}                          & \multicolumn{1}{c|}{} \\ 
\hline\hline
\xmark & \xmark & 0.9214 & 13\\
\hline
\xmark & \cmark & 0.9214 & 16\\
\hline
\cmark & \xmark & 0.9193 & 38\\
\hline
\cmark & \cmark & 0.9193 & 63\\
\hline
\end{tabular}
\end{center}

        \label{ta:opt}
    \end{table}
    
    
    
\begin{figure*}[h]
     \centering
     \begin{subfigure}[b]{0.16\linewidth}
         \centering
         \includegraphics[width=.95\linewidth]{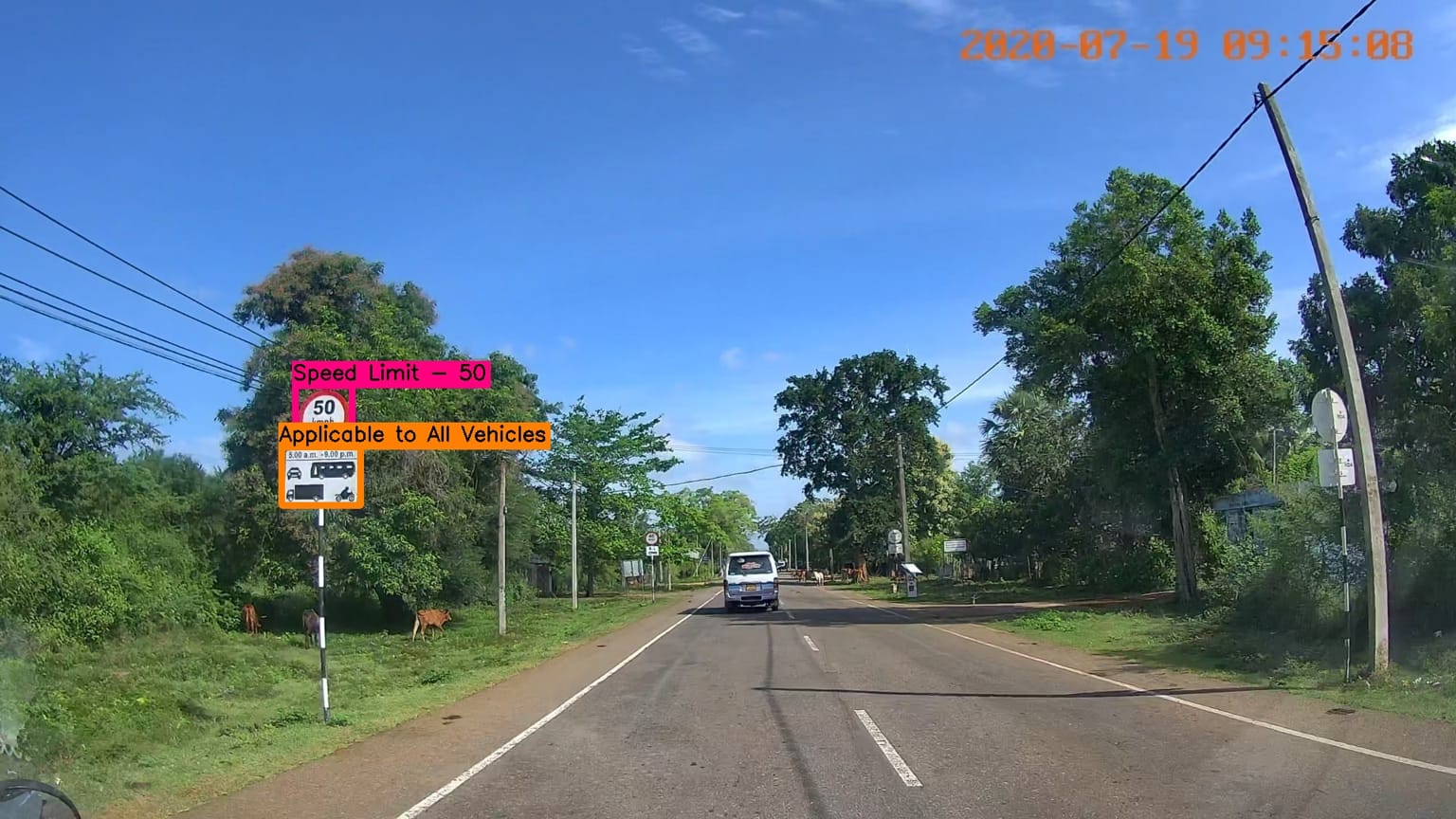}
     \end{subfigure}%
     \begin{subfigure}[b]{0.16\linewidth}
         \centering
         \includegraphics[width=.95\linewidth]{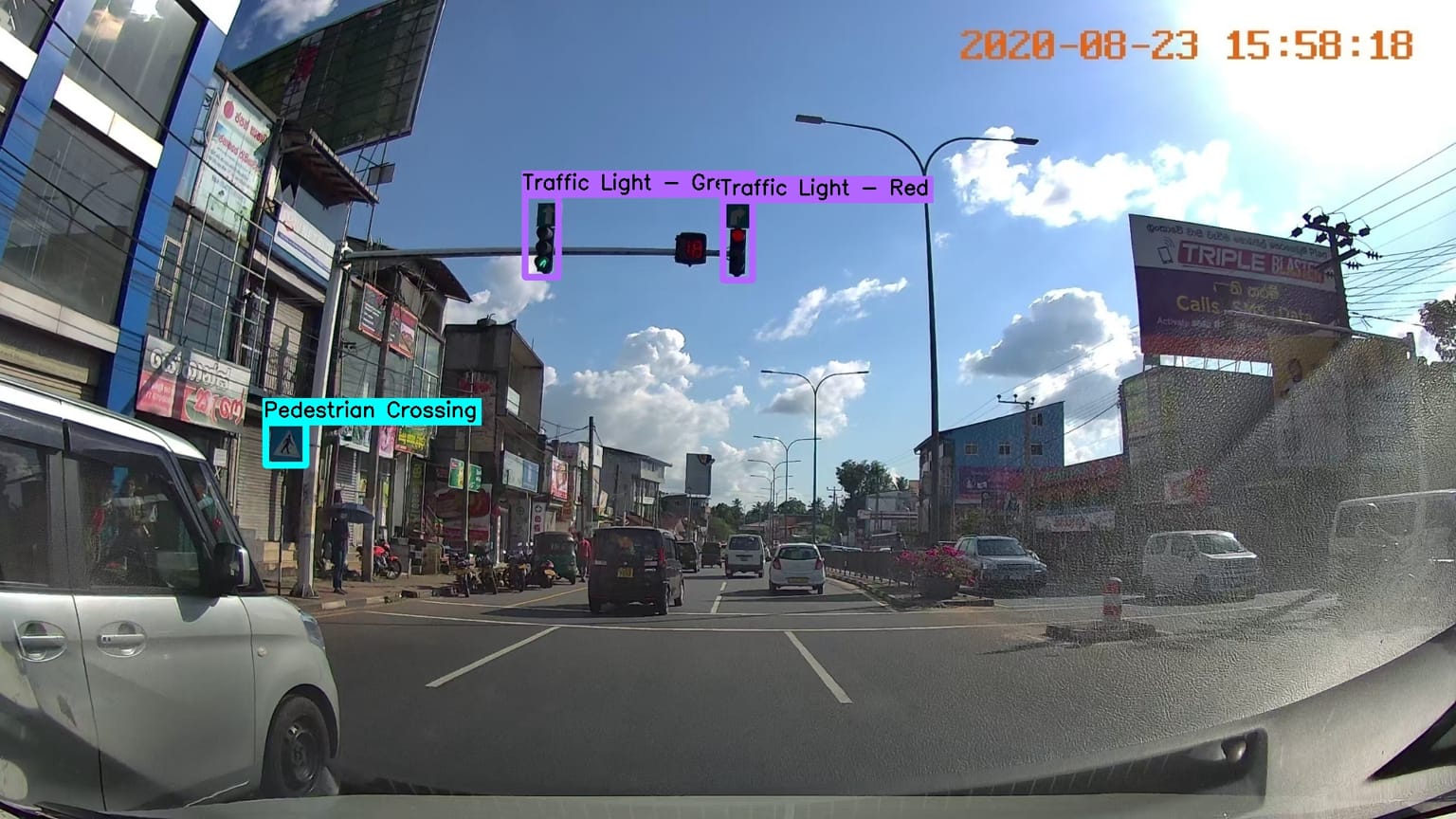}
     \end{subfigure}%
     \begin{subfigure}[b]{0.16\linewidth}
         \centering
         \includegraphics[width=.95\linewidth]{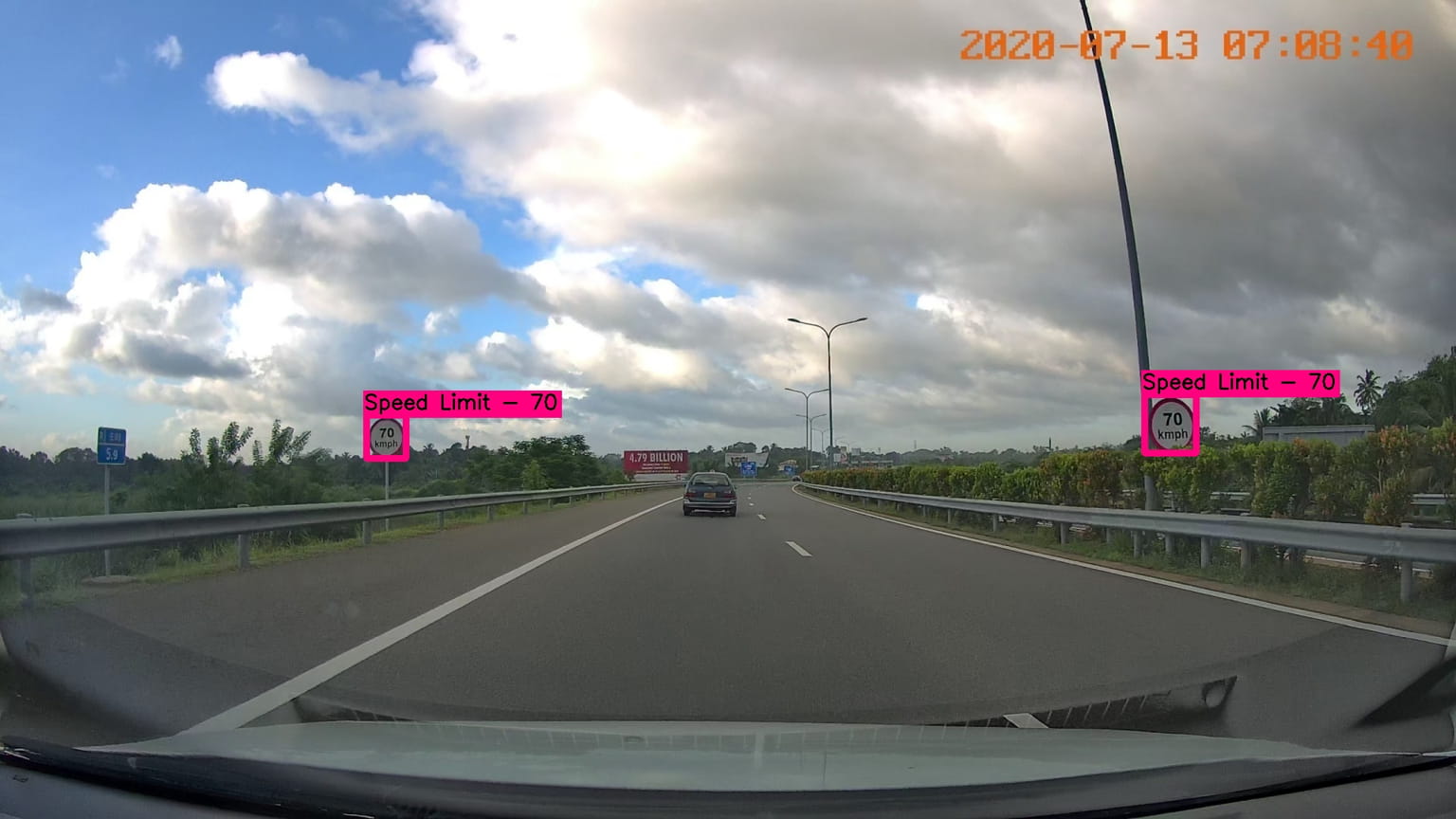}
     \end{subfigure}%
     \begin{subfigure}[b]{0.16\linewidth}
         \centering
         \includegraphics[width=.95\linewidth]{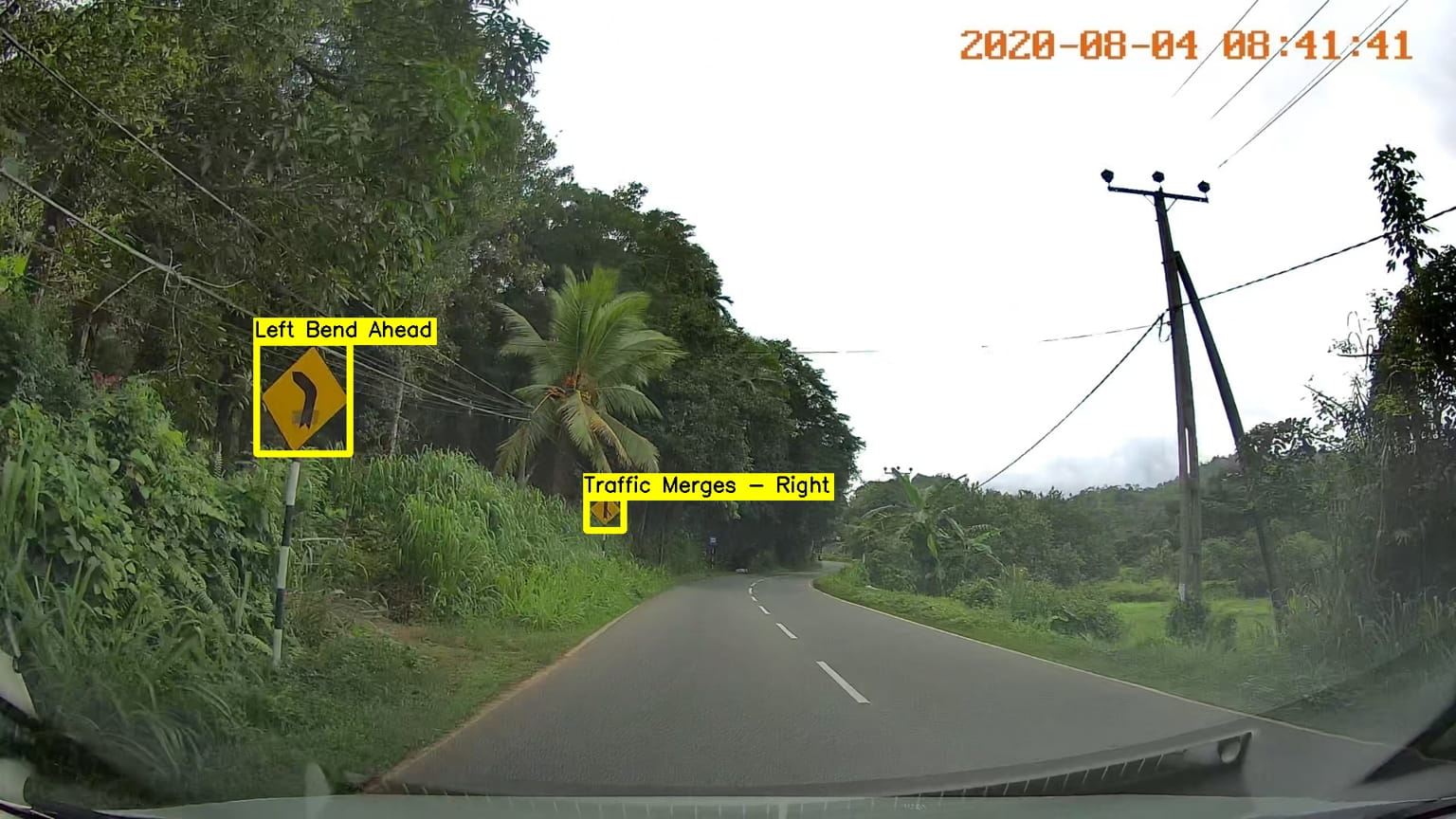}
     \end{subfigure}%
     \begin{subfigure}[b]{0.16\linewidth}
         \centering
         \includegraphics[width=.95\linewidth]{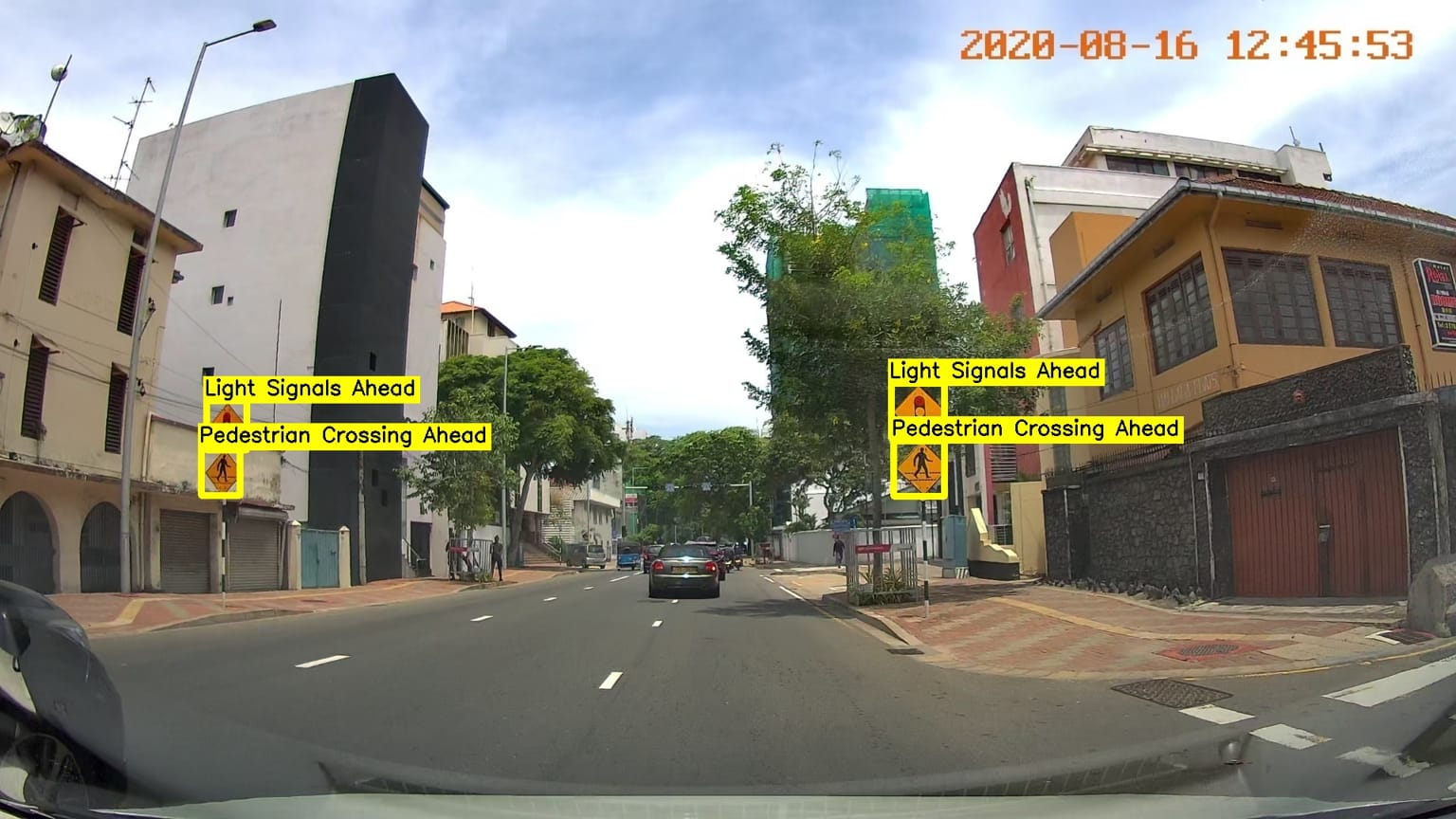}
     \end{subfigure}%
     \begin{subfigure}[b]{0.16\linewidth}
         \centering
         \includegraphics[width=.95\linewidth]{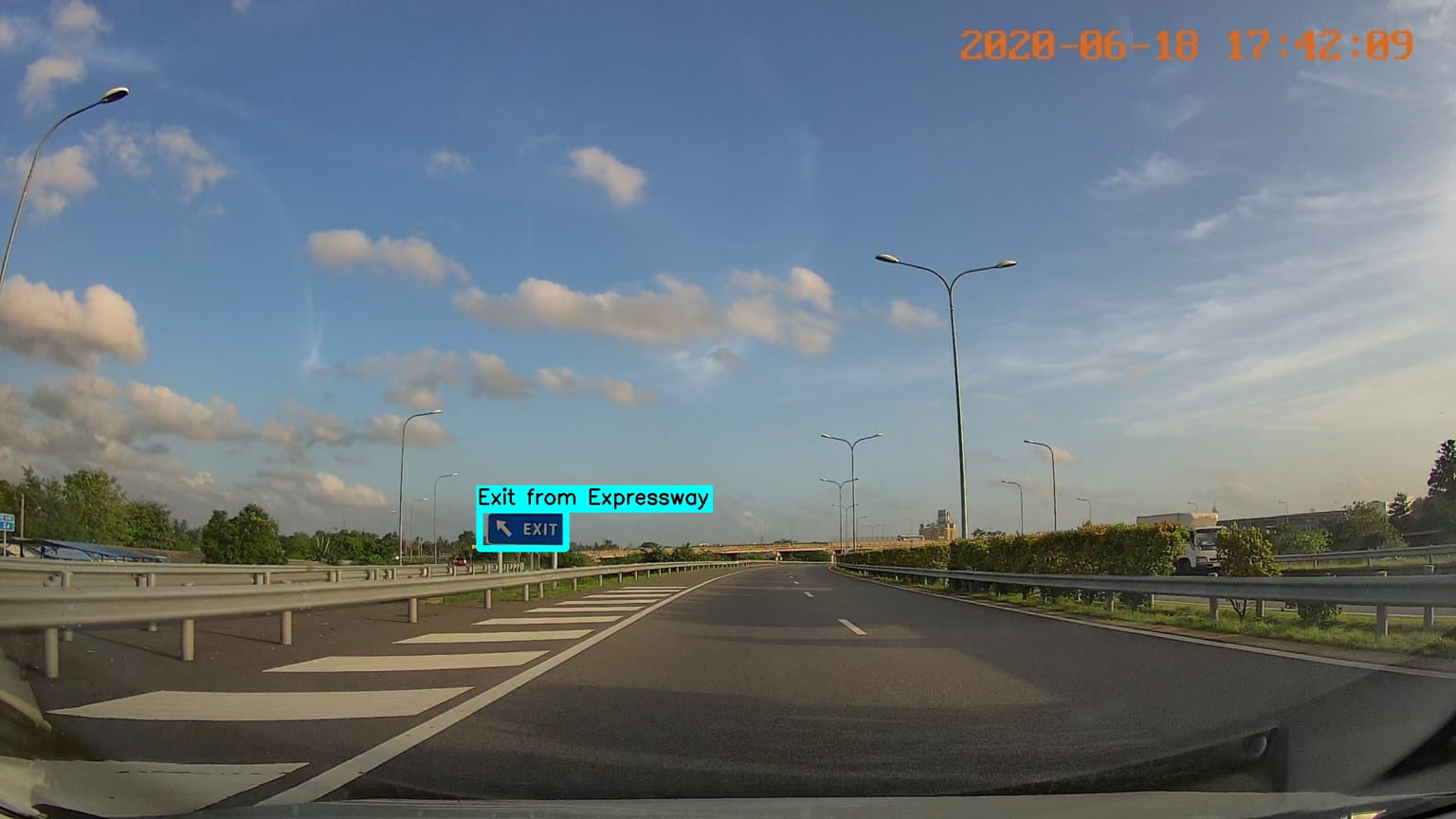}
     \end{subfigure}%
     
         \vspace{0.3em}
     
          \begin{subfigure}[b]{0.16\linewidth}
         \centering
         \includegraphics[width=.95\linewidth]{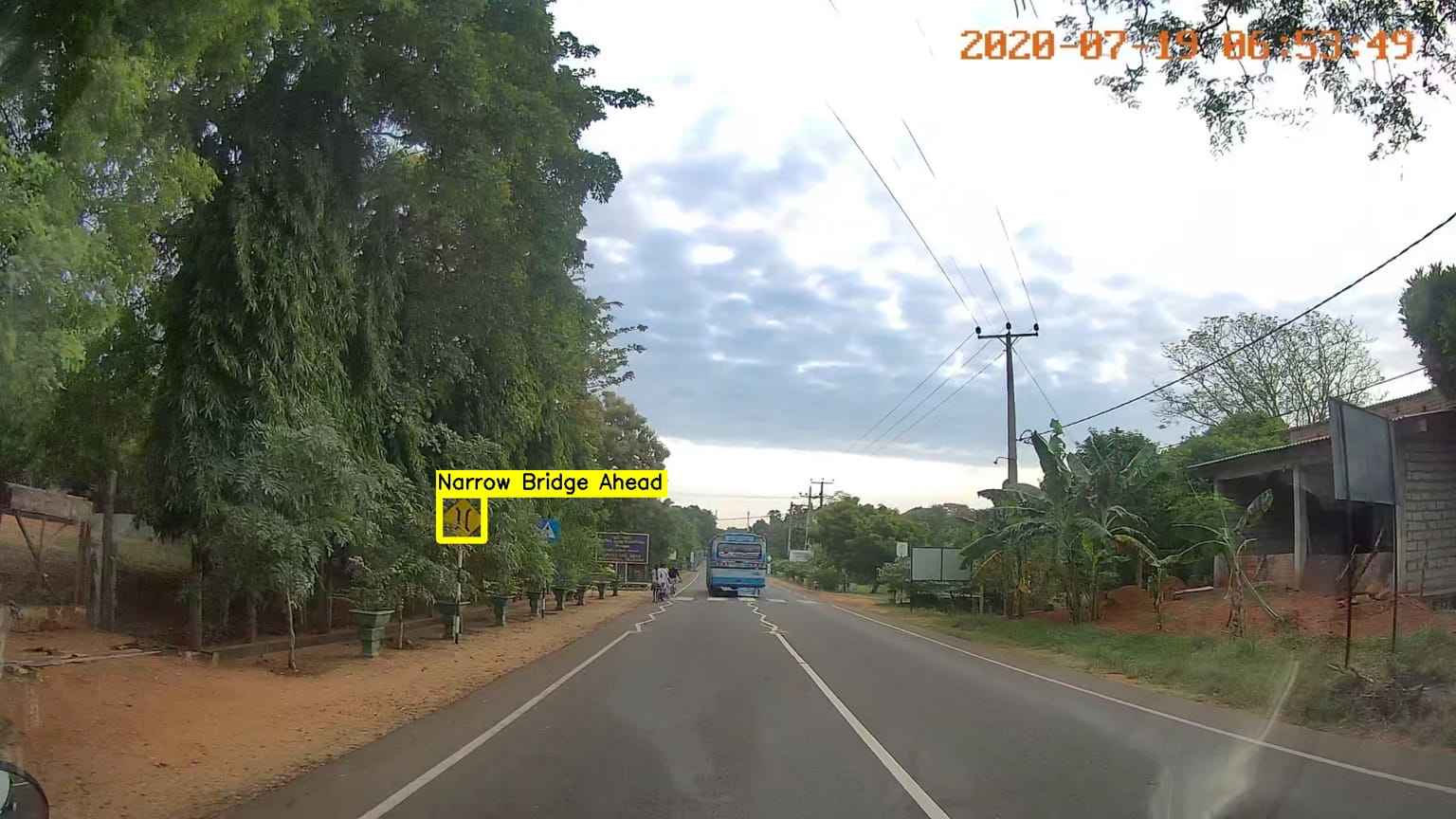}
     \end{subfigure}%
     \begin{subfigure}[b]{0.16\linewidth}
         \centering
         \includegraphics[width=.95\linewidth]{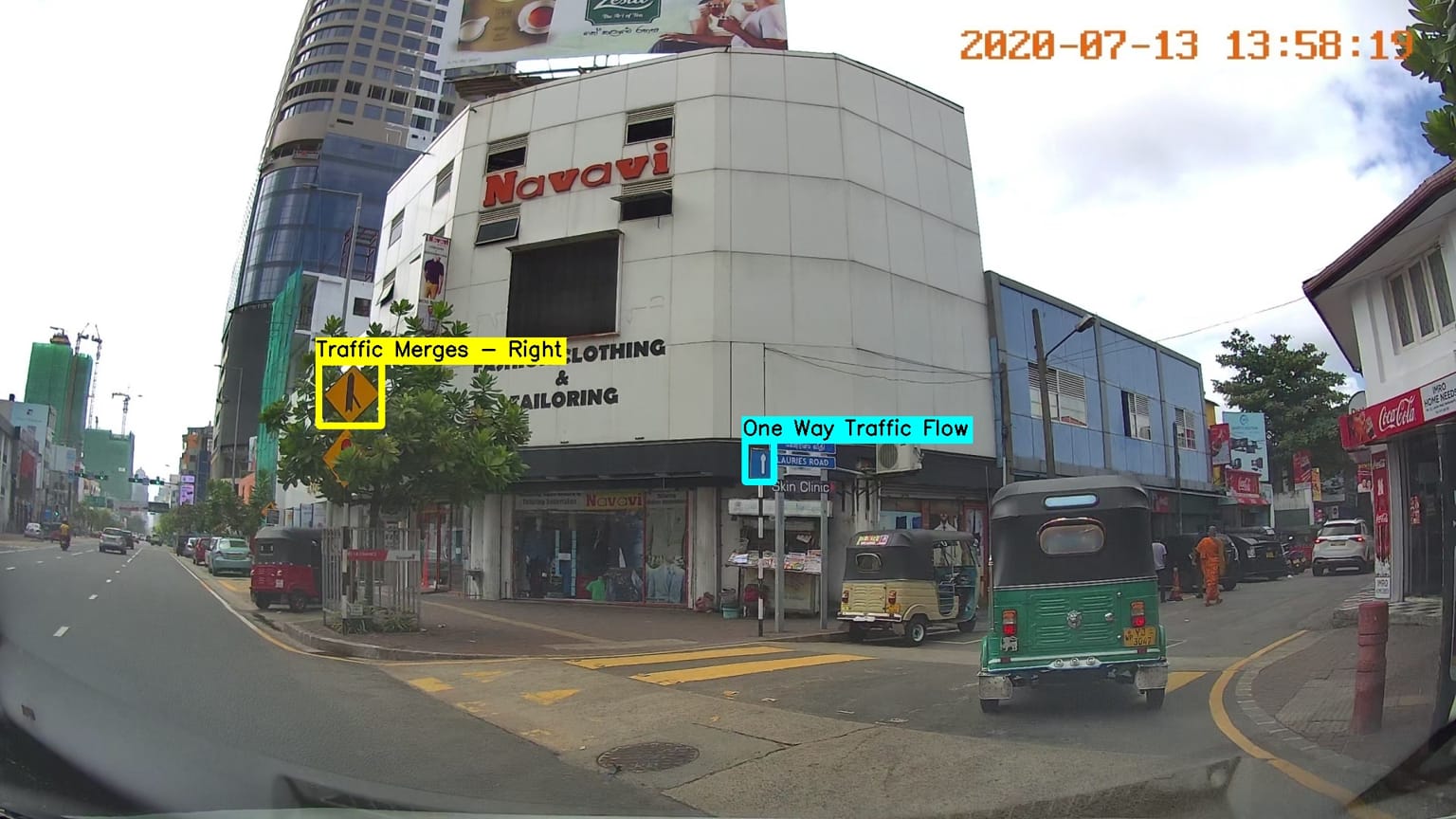}
     \end{subfigure}%
     \begin{subfigure}[b]{0.16\linewidth}
         \centering
         \includegraphics[width=.95\linewidth]{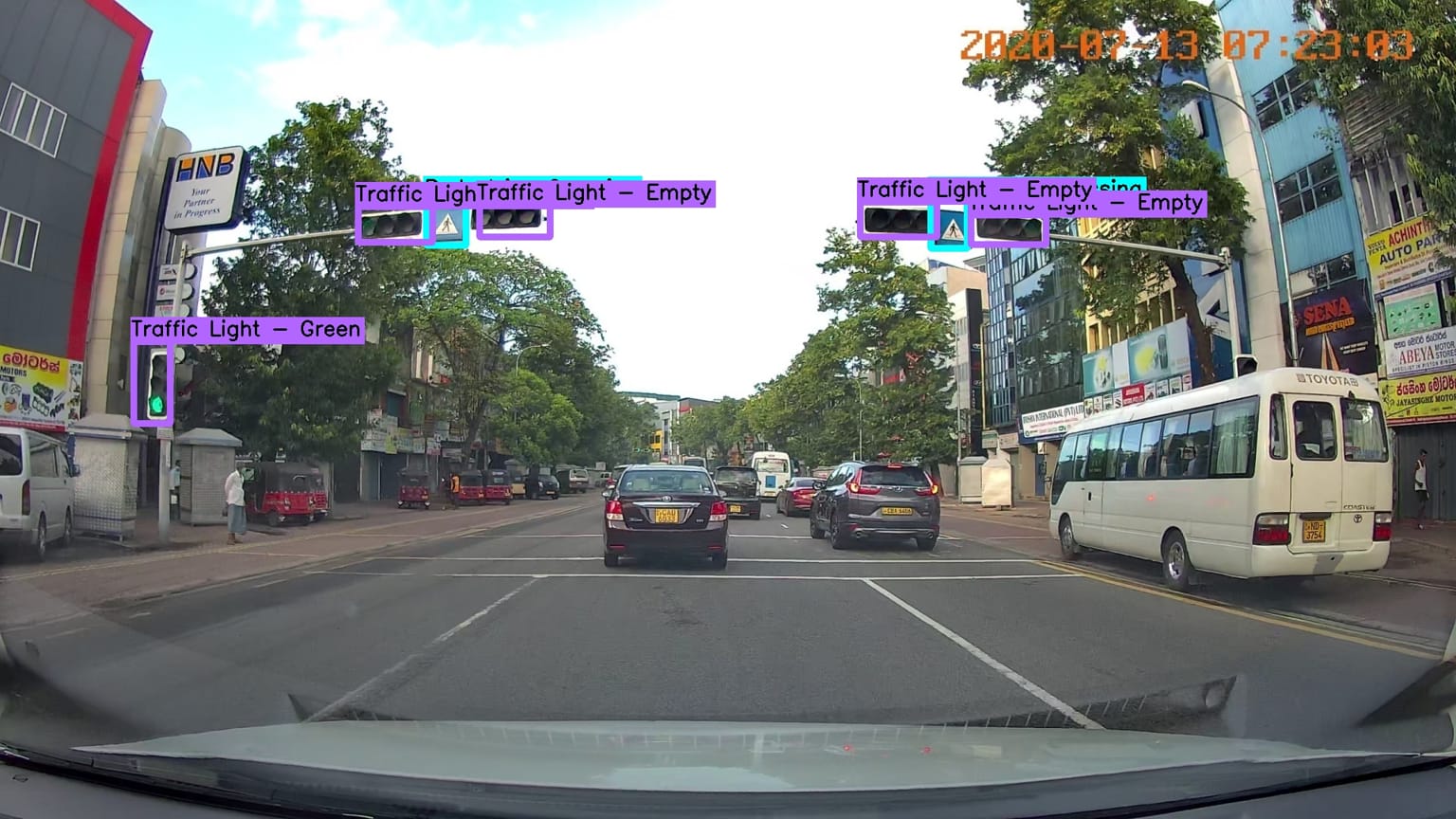}
     \end{subfigure}%
     \begin{subfigure}[b]{0.16\linewidth}
         \centering
         \includegraphics[width=.95\linewidth]{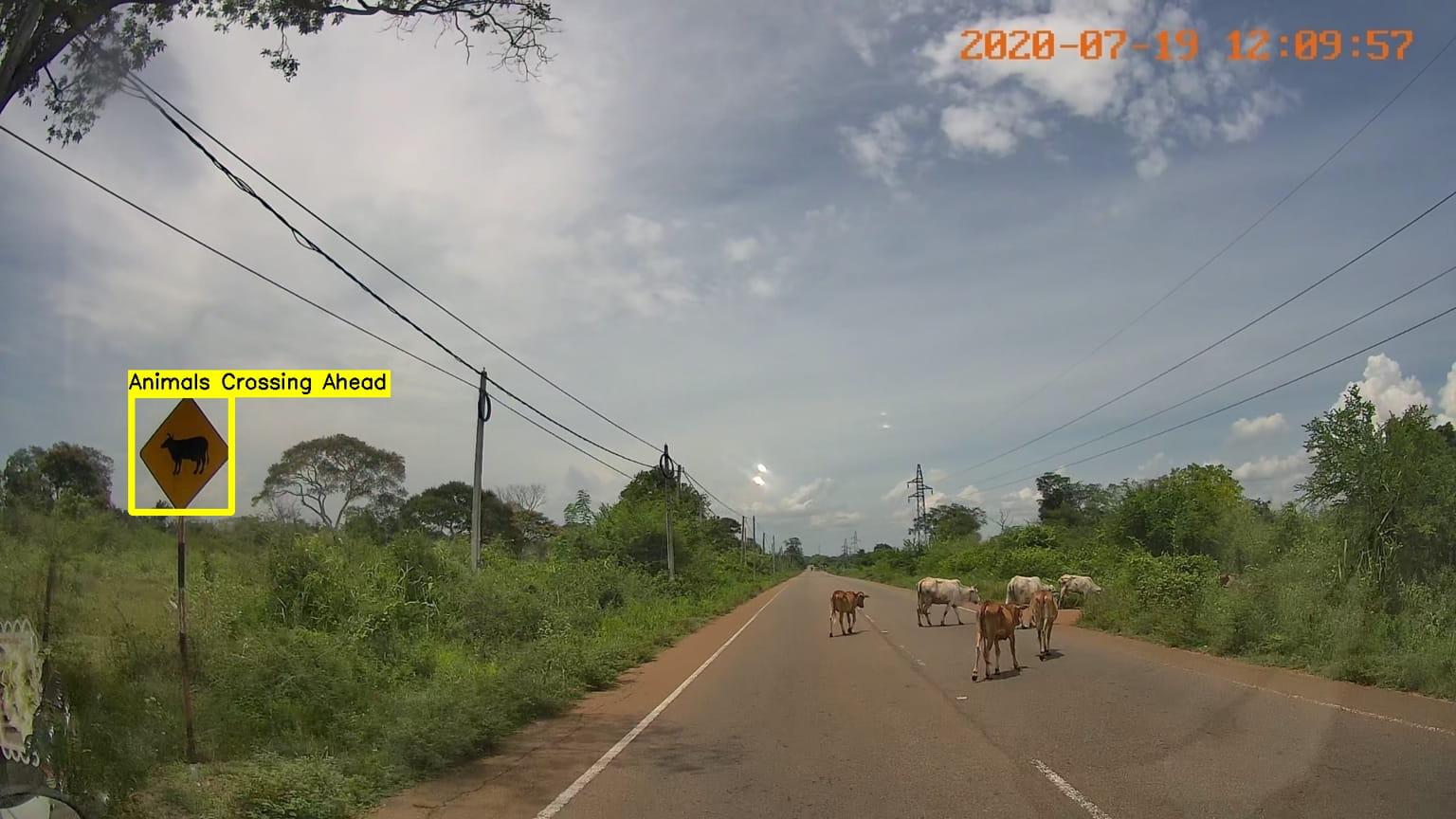}
     \end{subfigure}%
     \begin{subfigure}[b]{0.16\linewidth}
         \centering
         \includegraphics[width=.95\linewidth]{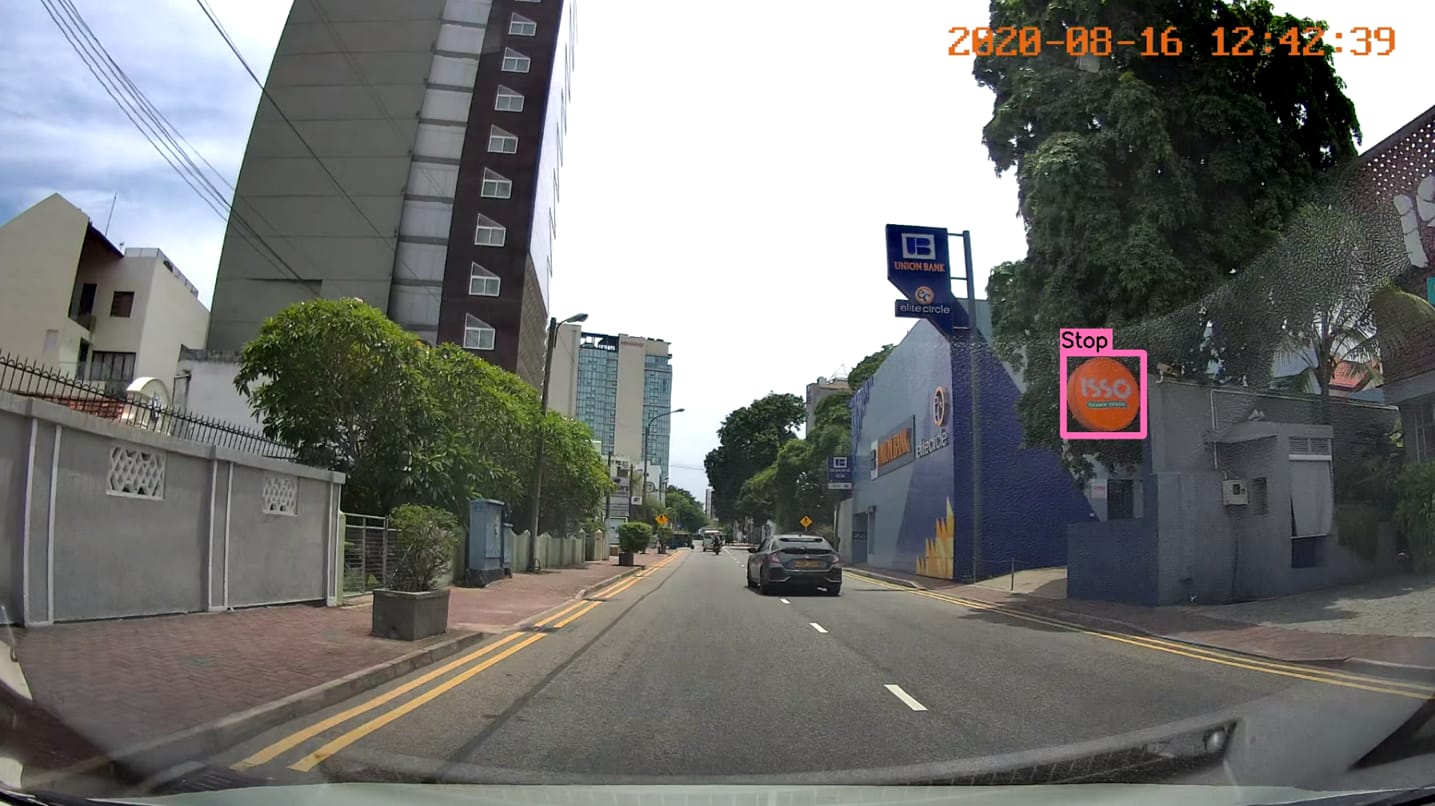}
     \end{subfigure}%
     \begin{subfigure}[b]{0.16\linewidth}
         \centering
         \includegraphics[width=.95\linewidth]{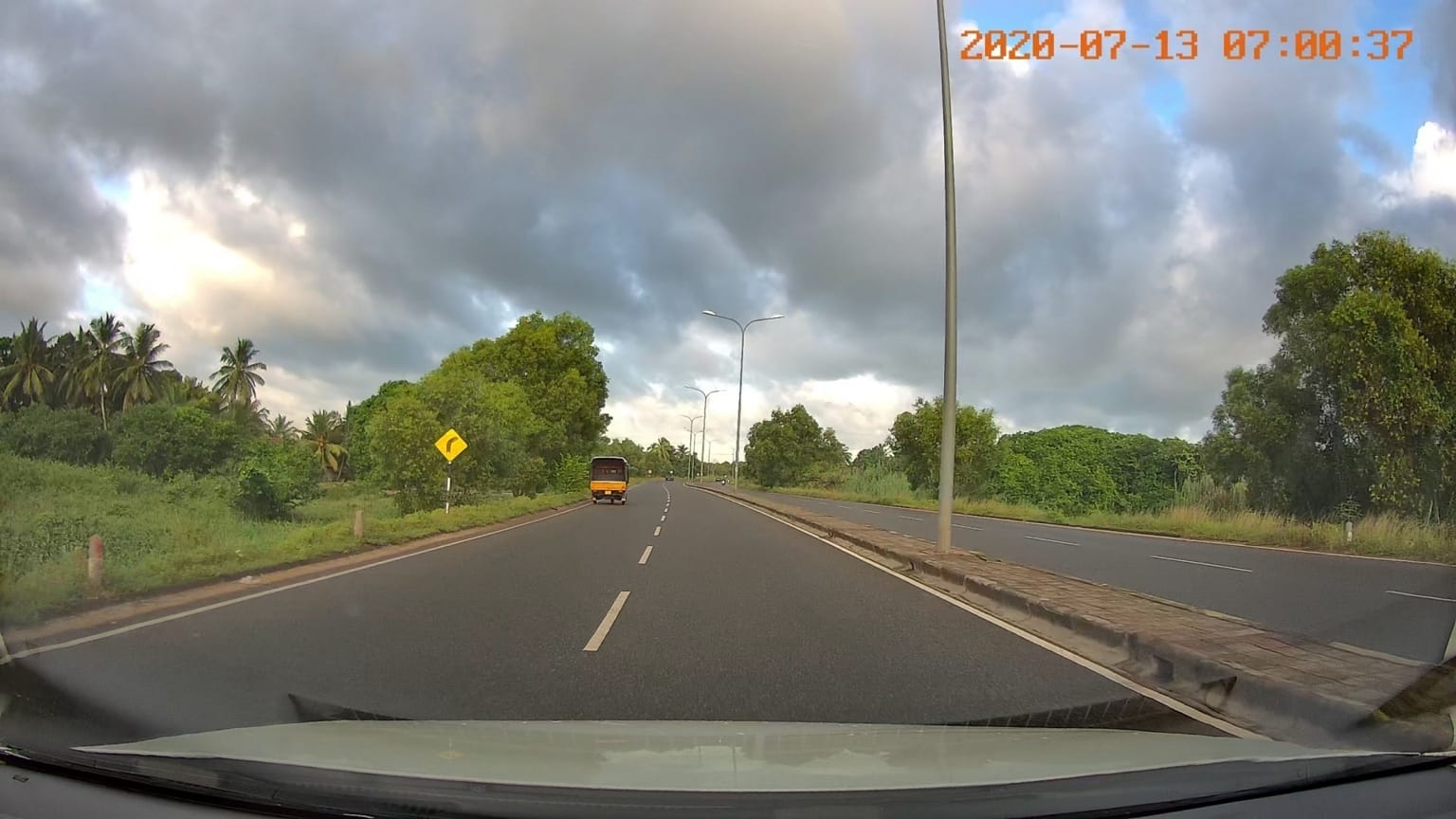}
     \end{subfigure}%

     
     \caption{ 
    Visualization of traffic sign and traffic light detection results. The first ten images show accurate detections in different road scenarios while the last two images show failure cases including false detections and undetected instances.}
     \label{fi:pictorial_results_ts}
\end{figure*}

    Some of the qualitative results of the traffic sign and traffic light detection task obtained by the SSD-MobileNet-v2 \cite{SSD,sandler2018mobilenetv2} model are visualized in Fig. \ref{fi:pictorial_results_ts} including urban, rural, expressway, dazzle light and occlusion conditions. Examples for false detections and undetected instances have also been included.
    
    \vspace{0.17cm}
    
    \section{Conclusion}
    
    
    In this work, we proposed a simple, end-to-end, deep learning based two-staged detection pipeline for real-time traffic sign and traffic light detection in an embedded system. Furthermore, we introduced the CeyRo traffic sign and traffic light dataset covering a wide range of challenging road scenarios in Sri Lanka. Our benchmark contains 7984 total images and 10176 traffic sign and traffic light instances belonging to 70 traffic sign classes and 5 traffic light classes. The effectiveness of the proposed framework is justified using both qualitative and quantitative results. We further demonstrated the capability of our system to deliver real-time performance in an embedded system using an Nvidia Jetson AGX Xavier device. The detection models were optimized using TensorRT and integrated with Robot Operating System to deploy as a traffic sign and traffic light detection system which achieves a high inference speed of 63 FPS. We believe this is a promising step towards real-time traffic sign and traffic light detection in challenging road scenarios with limited computational resources.

    \newpage
    
    {\small
    \bibliographystyle{ieee_fullname}
    \bibliography{egbib}
    }
    
    \end{document}